\pdfoutput=1
\documentclass[10pt,final,conference,letterpaper,oneside,twocolumn]{IEEEtran}

\usepackage[cmex10]{amsmath}
\usepackage{graphicx}
\usepackage{url}
\usepackage{xspace}

\graphicspath{{images/}}
\DeclareGraphicsExtensions{.pdf,.jpeg,.png}

\newcommand{\nop}{NimbRo\protect\nobreakdash-OP\xspace}
\newcommand{\nimbro}{NimbRo\xspace}
\newcommand{\scl}{State Controller Library\xspace}
\newcommand{\scls}{SC Library\xspace}
\newcommand{\bcf}{Behavior Control Framework\xspace}
\newcommand{\bcfs}{BC Framework\xspace}
\newcommand{\secref}[1]{Section~\ref{sec:#1}\xspace}
\newcommand{\figref}[1]{Figure~\ref{fig:#1}\xspace}

\newcommand{\chapref}[1]{Section~\ref{chap:#1}\xspace}
\newcommand{\cpp}{C\texttt{\nolinebreak\hspace{-.05em}+\nolinebreak\hspace{-.05em}+}\xspace}

\usepackage{eso-pic}

\AtBeginDocument{\AddToShipoutPictureFG*{\AtTextUpperLeft{\put(0,\LenToUnit{21pt}){\parbox{\textwidth}{\centering\bfseries
Proceedings of 8th Workshop on Humanoid Soccer Robots, International Conference on\\Humanoid Robots (Humanoids), Atlanta, USA, 2013
}}}}}
\begin{document}

\bstctlcite{IEEEexample:BSTcontrol}

\title{Hierarchical and State-based Architectures for Robot Behavior Planning and Control}

\author{\vspace{-3em} \\
\IEEEauthorblockN{Philipp Allgeuer and Sven Behnke\vspace{0.8em}}
\IEEEauthorblockA{Autonomous Intelligent Systems, Computer Science Institute VI, University of Bonn, Germany\\
\tt\small pallgeuer@ais.uni-bonn.de, behnke@cs.uni-bonn.de\\
http://ais.uni-bonn.de}}

\maketitle

\begin{abstract}
In this paper, two behavior control architectures for autonomous agents in the
form of cross-platform \cpp frameworks are presented, the \scl and the \bcf.
While the former is state-based and generalizes the notion of states and finite
state machines to allow for multi-action planning, the latter is behavior-based
and exploits a hierarchical structure and the concept of inhibitions to allow
for dynamic transitioning. The two frameworks have completely independent
implementations, but can be used effectively in tandem to solve behavior
control problems on all levels of granularity. Both frameworks have been used to
control the \nop, a humanoid soccer robot developed by team \nimbro of the
University of Bonn.
\end{abstract}

\section{Introduction}
\label{chap:intro}

The programming of complex agent behaviors that are intended to function in
highly dynamic environments can be a challenging task in modern robotics.
Without the support and uniformity of a well-defined behavior control
architecture, this task can be even harder. Two such behavior control
architectures, implemented as cross-platform \cpp frameworks, are presented in
this paper and proposed for use. These are the \scl (\scls) and the \bcf
(\bcfs).

An agent system with clearly defined inputs and outputs is considered, one that
at every instant must select its output action(s) based on its signal inputs.
These signals may involve sensory inputs that quantify characteristics of the
environment, and/or signals directly from external sources or other agents. This
action selection process is referred to as the \emph{behavior control} problem,
where a behavior is taken to be an observable and coordinated pattern of
activity of an agent, involving action and/or response to stimuli from the
environment \cite{BehaviourDefn}. Every robotic agent requires some form of
architecture for behavior control in order to complete the tasks that it is
given. Simple reactive feedback control loops have been used since the earlier
days of robotics with much success, but with the increasing complexity of tasks
and capabilities of robotic systems, these have grown insufficient to be a
complete solution for applications such as robot soccer and domestic service
robotics.

The challenge is to find an efficient and modular way of representing and
programmatically implementing complex behavioral systems in code, ideally in
the form of a library or framework. Any such construct needs to facilitate the
implementation and encapsulation of near-arbitrary control systems, capable of
dealing with real-world effects such as disturbances, sensor noise, environment
stochasticity, and partial controllability and/or observability. These are
conditions for which traditional artificial intelligence (AI) architectures were
generally not designed to handle, with simplified and abstracted virtual
environments being more typical platforms for the development of AI than the
highly dynamic environments generally encountered by autonomous mobile agents. A
trivial example of this would be the observation that, while an agent that simply
uses a predetermined sequence of actions for its behavior may be
consistently successful in a structured and deterministic software environment,
a similarly controlled robot would likely fail whenever any of the
aforementioned real-world conditions come into play. In addition to the efficiency
and modularity requirements of a behavior framework, it is also desirable for
such a framework to inherently support the notion of planning into the future,
as opposed to focusing on purely reactive control. This is specifically
addressed by the \scls, as described later in \chapref{scl}.

Both frameworks were designed, implemented, utilized and tested within the
setting of humanoid robot soccer. As opposed to being mutually exclusive for an
application, these two frameworks were written in a way that they can easily and
effectively be used in tandem. The \nop robot \cite{NimbroOPDescription13},
developed by team \nimbro of the University of Bonn, was used as a testbed. The
\nop is a humanoid platform that is open source, both in terms of its hardware
and software. As such, the \scls and the \bcfs are available as part of the
\nimbro Robot Operating System (ROS) soccer package software release for the
\nop. It is to be noted however that despite their origins in the area of
humanoid robot soccer, the frameworks were written to be completely generic,
allowing them to be used in virtually any type of robotic system---or even
software system---that requires some notion of behavior control. Nevertheless,
in a broad sense, the two frameworks are targeted for use in real-time systems.
The \nop currently makes use of the \bcfs for its soccer behaviors.

\section{Related Work}
\label{chap:relatedwork}

As behavior control is one of the fundamental problems in the field of robotics,
many approaches have been developed in the past. Behavior control has been
heavily researched within the context of artificial intelligence
\cite{ArtificialIntelligence}. This has led to a number of architectures and
classical artificial intelligence approaches. These were generally developed
within the setting of simplified virtual environments however, and so were not
designed for use on robots in highly dynamic real-world environments, with all
the nonidealities that come with it. An example of such an architecture is the
Belief, Desire and Intention (BDI) agent model \cite{Rao91modelingrational},
\cite{Rao95bdiagents}. This model is based on modal logics and the partial or
complete axiomatization thereof. Multiple BDI logic variants exist, but all of
them in general include modalities for beliefs, desires, intentions,
capabilities, actions, agency, and time \cite{Rao95formalmodels}. A strength of the
BDI logics is their strong formalization and theoretical foundations, but it
is a non-trivial task to apply such logics to describe and control robots in
real-world applications \cite{Burkhard1998}.

Earlier approaches to behavior control from the field of robotics include the
\emph{subsumption architecture}, a behavior-based approach. This was originally
proposed by Brooks in \cite{Brooks86Orig}, but was later modified in
\cite{Brooks89} and \cite{Connell89}. The idea of the subsumption architecture
is to arrange the behaviors in a layer hierarchy where higher layers can
influence and suppress the data flow of lower layers in order to achieve higher-level
goals. Brooks constructed a number of robots to demonstrate the
architecture, leading in particular to robots with insect-level behavior and
intelligence \cite{Brooks91Ex}, \cite{Maes1991ANA}. Adaptation of the
architecture to more complex systems proved to be difficult.

Maes developed the Agent Network Architecture (ANA) \cite{Maes1991ANA},
\cite{Maes1990Sit}, also a behavior-based artificial intelligence approach. In
this architecture, the agent consists of a distributed and decentralized
collection of primitive behaviors, referred to as \emph{competence modules}.
The competence modules are divided into groups of incompatible behaviors, and
interact based on a network of predecessor, successor and conflictor links in
such a way that the various modules are activated and inhibited dynamically when
appropriate. Each competence module only implements a single basic primitive
behavior however, so the activation network would grow quite large for
real-size problems and become slow and overwhelmed by details
\cite{Maes1991ANA}. Also, it is a difficult task to tune all of the network
parameters and activation functions so that the appropriate module(s) for a
given situation in general tend to be activated.

In a different category to the previously mentioned approaches to behavior
control are the class of behavior control languages---languages that were
specifically designed for the specification of behaviors, usually on a rather
conceptual level. Examples include the Behavior Language by Brooks
\cite{BehaviourLanguage}, which was based on \cite{Connell89} and his prior work
on the subsumption architecture, the Configuration Description Language
\cite{MacKenzie1997Thesis}, and Colbert \cite{Colbert97}, part of the Saphira
Control Architecture \cite{Saphira98}. A more recent example of a behavior
control language is the \emph{Extensible Agent Behavior Specification Language
(XABSL)}. Originally developed in \cite{XABSLThesis04} as an XML-based language,
it has been extended and improved incrementally in works such as
\cite{XABSLThesis09}, which introduced a behavior language representation for
XABSL with a more compact syntax. The idea behind XABSL is to use a hierarchy of
finite state machines called \emph{options} to select the appropriate action(s)
to execute from a set of basic behaviors. XABSL currently finds relatively
widespread use in humanoid soccer.

Behavior languages in general excel at abstracting away coding particularities
of more generic low-level programming languages such as \cpp, allowing the main
focus of behavior coding to remain more on `what' rather than `how'. This also
generally allows more succinct representations of behaviors, or the interactions
between them, to be constructed. The main disadvantage of using a behavior
language is the overhead of coupling two different programming languages
together, in terms of both project architecture and runtime considerations. A
runtime engine is often required to execute the resulting behaviors and
integrate them with the remaining code, which is not as efficient as behavior
implementations that are seamlessly integrated using the one target language.
The interfacing of data signals between the two languages can also be a
challenge, often limiting the flexibility of such behavior control approaches.
Neither of the frameworks presented in this paper is a language in its own
right.

The next category of behavior control architectures are the state-based
techniques. An example is the hierarchical control structure for mobile agents
proposed in \cite{hierarchicalFSMs}. As the behaviors of a robot under
development often grow incrementally from rather simple beginnings, it is not
uncommon for a behavior control system to consist simply of a custom
implementation of a finite state machine. When this becomes too cumbersome, or
is otherwise not desired, many finite state machine code libraries exist that
can be used to implement such basic state-based approaches (e.g.
\cite{MachineObjects}). To the knowledge of the authors however, no other simple
state machine implementation exists that allows for the planning of multiple
future states as the \scl presented in this paper does. This was a feature that
was considered to be important for flexibility and extensibility reasons.

\section{The State Controller Library}
\label{chap:scl}

\subsection{Overview}
\label{sec:scloverview}

\begin{figure}
\centering
\includegraphics{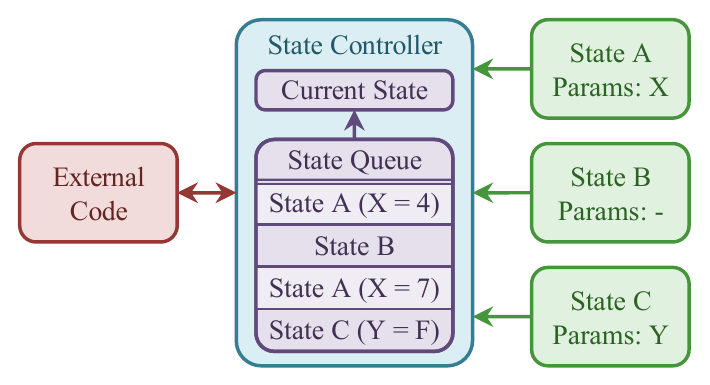}
\caption{A block diagram of the \scls architecture with three sample states and
a populated state queue.}
\label{fig:sclarch}
\end{figure}

The \scl is a generic platform independent \cpp framework that allows finite
state machines, hierarchical state machines, and multi-action planning
generalizations thereof to be realized. The structure and implementation of the
library focuses on the application of such architectural constructs to real-time
control loops, but can be reasonably adapted for virtually any other
application, even completely unrelated to control systems. While the underlying
ideas and structure of the \scls are of predominant interest here, it is worth
noting that the simplicity of this behavior control approach allows for a very
lightweight, unintrusive and resource-efficient implementation.

The core idea of the library is to have a state controller object with a certain
collection of state object types that are bound to it. These state object types
(i.e. classes in \cpp) are henceforth simply referred to as the \emph{states} of
the controller. Instantiations of a state are referred to as instances of that
state, or \emph{state instances}. These instances are executed by the state
controller in the required order, as stipulated by a dynamically maintained list
of desired future states. This list, which is embedded inside the corresponding
state controller, is referred to as the \emph{state queue}. Each state is
responsible at runtime for specifying its own outgoing state transitions by
modifying this state queue. Although it is not necessary for an application to
use the full flexibility of the queue---there are abstractions available in the
\scls that hide away the details of the queue if only simple functionality is
required---it is still available at little to no performance cost. This
increases the extensibility of control systems implemented using the library,
and caters for applications where some form of action planning is desired. An
overview of the \scls architecture is shown in \figref{sclarch}.
The latest release of the library as at September 2013 is version 1.2, and is
available for download from the project page \cite{SCLSourceforge},
where all future releases will also be made available.

\subsection{Motivation}
\label{sec:sclmotivation}

The conception of the \scl was driven by two main considerations. The first was
the need for a simple first solution to the behavior control problem for the
\nop robot. When the \nop platform was first being programmed with basic soccer
skills, a simple state machine framework was required to write the behaviors,
pending a more powerful and long-term solution. In anticipation of the creation
of the more powerful behavior control framework however, it was a second core
consideration that the \scls be able to be used to implement finite state machines within the individual
behaviors of the future framework (i.e. the \bcfs). This would be for the case
that an individual behavior requires a notion of sequentiality and state that
is best implemented locally. An example of this is a walking or kicking
behavior that uses different states internally for the different phases of the
corresponding motion. As such it was important that the \scls be able to be used
for applications ranging all the way from the implementation of a whole soccer
behavior system, down to the implementation of the smallest and most
fundamental finite state machines in the code---without incurring any
unnecessary overhead or performance losses. The performance and efficiency of
the library was of particular importance due to the real-time nature of the \nop
control task.

Further attributes that were desired of the \scls were for it to be robust to
incorrect use, able to plan multiple actions into the future, and able to
integrate seamlessly with the target code of the \nop. The structure of the
implementation was also desired to be control-based, not event-based. No
existing state machine library was found that addressed all of the
requirements outlined in this section.

\subsection{Library Structure}
\label{sec:sclstructure}

The \scl is written in \cpp, so it is able to make use of the object-oriented
programming paradigm to break down the state control structure into a collection
of objects. As previously indicated, the main objects in use are the \emph{state
controller}, the \emph{state queue}, and the individual \emph{state instances}.
These terms were carefully defined in \secref{scloverview}.

At the core of the \scls is a step routine that controls the execution of the
state controller. The routine may be called continuously or at arbitrary
intervals, but the intended use for control applications is for it to be called
from a timed loop running at some nominal rate. Each step executes one so-called
cycle of the state controller. In each cycle, it is first checked whether
the currently executing state instance has set a flag in the last cycle that it has
completed its task. If so, the state instance sitting at the head of the state
queue is popped from the queue, activated and executed. If this
state instance also completes its task some number of cycles later, during which
time it is executed once per step, it is deactivated before the successor
state is activated. Users of the library have the option to write specific code
to handle the activation and deactivation events of the individual states. Once
a state instance has been deactivated, it is marked as complete and can never be
executed again. To reenter a previous state, a new instance of that state simply
needs to be created and placed into the queue, as normal. Callbacks are made
available to the user at all the stages of the main loop so that
application specific code can be injected as necessary.

The motivation behind having individual state instances intended for single use
only is so that multiple instances of a state can exist concurrently within the
state queue, and so that the various state instances can be individually
configured via \emph{state parameters}. These are parameters that are passed to
a state instance at construction time, and are used to specialize the task of a
particular instance. For example, if a state has the task of bringing a robot to
a particular global pose, then the state parameters can be used to specify the
target global pose for a particular state instance. The power of this method is
that multiple instances of a state within the queue can have differing
customized objectives. To achieve even comparable results, standard finite state
machines would somehow have to keep track at all times of where in a desired
sequence of actions the currently executing state fits in, and modify the state
objectives locally. This is not only more complex and error-prone to implement,
but it also forces lower-level states such as locomotion states to be imparted
with knowledge of higher-level goals.

\subsection{State Transitions and the State Queue}
\label{sec:scltransitions}

Each state instance when it executes is responsible for
modifying the state queue, the dynamically manipulatable ordered list of states
that are pending execution by the state controller. As would generally occur
when there are multiple items in the queue though, a state is not actually
obliged to modify it. This allows low-level behavior states to be
implemented (such as for locomotion) that do not need to deal with any knowledge
of higher-level planning, as they do not need to specify their successor state. The
successor state is already uniquely determined as the state that was previously
placed at the head of the queue. As such, the higher-level planning can
naturally be separated from the lower-level states that actually execute the plans,
reducing the code complexity and clarifying its programmatic and behavioral
intent compared to what would be expected with more traditional finite state
machine implementations. It is important to note however, that an agent is still
not committed to a plan once it has been pushed into the state queue. The queue
can still be modified at any time by clearing, inserting, removing, rearranging,
etc.\ the states as required, which can be necessary if significant changes in
the environment are perceived.

For smaller applications of the \scls, pure simple next-state logic can be
achieved by configuring the individual state instances to only ever add a single
item to the end of the state queue, guaranteeing that this item will be at the
head of the queue and the one and only successor state. Convenient shortcuts for
this common operation exist within the library. In larger applications of the
\scls where the number of possible transitions grows quadratically with the
number of states, it often occurs that certain groups of states have a
collection of similar outgoing transitions, activated on similar preconditions. To
avoid unnecessary code duplication, generic transition routines can be embedded
inside the state controller object to apply the appropriate collection of
outgoing transitions to multiple states at a time. Taking the idea of grouping
the states a step further leads to a natural extension of the \scls---if a parent
state controller is used to selectively switch between a set of subordinate
state controllers that each implement one of the groups of the original states,
then a hierarchical state machine structure like the one in
\cite{hierarchicalFSMs} emerges. It is then possible to go even further and take
advantage of the state queues embedded in each of the subordinate controllers,
allowing for even greater flexibility.

A fundamental difference between the \scls and a large proportion of the other
state machine libraries that already exist is that no explicit definition of a
transition map or transition table is required in the code. An example of an
entry in such a transition table would be a rule specifying that the combination
of a certain event occurring in a certain state should cause a transition to a
particular successor state. When employing simple next-state logic using the
\scls, this is encapsulated in the executed target code of a state, when a
transition to another state is triggered from within a conditional expression.
The `event' in this case would be the change of the evaluation of the
conditional expression from false to true. No direct analog of a transition
table exists however when the state transitions come about due to the sequencing
of multiple states in the state queue (although this is still equivalently performed by
each of the individual state instances). This is regarded as a feature though, as
it allows for more flexible and dynamic transitioning behavior, often desired
in real-world applications.

\subsection{Example}
\label{sec:sclexample}

\begin{figure}
\centering
\includegraphics{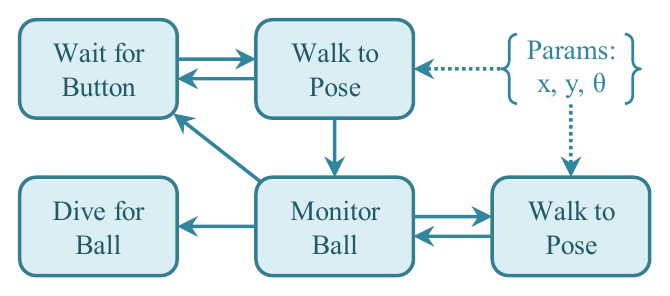}
\caption{A simplified example of a goalie behavior using the \scls.
Boxes represent the state instances, and the arrows indicate the possible
transitions between them. The \emph{Walk to Pose} state has three state
parameters.}
\label{fig:simplegoalie}
\end{figure}

Due to the large number of states and possible transitions in a typical behavior
control system, a complete example for a humanoid soccer behavior controller is
not given here. Consider however the simplified goalie behavior controller
presented in \figref{simplegoalie}. The goalie starts at the sidelines and must
wait for a button press before it starts walking to its designated position in
the goal. It then monitors the location of the ball based on its visual
detections and can decide to dive at any time to protect the goal. While the
goalie is keeping track of the ball, it may also decide to walk to a better
defensive position within the goal area, before continuing to monitor the ball.
The two instances of the \emph{Walk to Pose} state have been kept separate in
the diagram as a conceptually different task is being performed in each case,
despite being the same underlying state. Also, as indicated in the figure, the
\emph{Walk to Pose} state has three state parameters, which are used to specify
the target pose of the robot.

When the robot is started and the state controller is initialized, three states
would immediately be pushed into the queue, \emph{Wait for Button},
\emph{Walk to Pose} and \emph{Monitor Ball}. This has the effect that
\emph{Walk to Pose} does not need to specify, or even know, that its
desired successor state in this case is \emph{Monitor Ball}, and allows for a
simpler and clearer implementation of the behavior. Once in the monitoring
state, if the goalie decides to walk to a better location,
it would enqueue a \emph{Walk to Pose} state, followed by another instance of
the \emph{Monitor Ball} state, and mark the current
state instance as being complete. This would once again relieve the \emph{Walk to
Pose} state of needing to know for what particular purpose, or as part of what
plan, it was called upon. It is emphasized however that the goalie can still
clear the queue and repopulate it at any time if the enqueued states are no
longer appropriate. For example, if the button is pressed while the goalie is
still walking out to its desired field position, the future monitoring state can
be cleared from the queue and replaced with an instance of the \emph{Wait for
Button} state. The use of a state queue would also be advantageous if for
example a fixed obstacle was detected on the way to the goal area. The robot
would be able to dynamically prepend extra \emph{Walk to Pose} states to the
front of the queue in order to avoid the obstacle, all the while not
`forgetting' the desired successor state on arrival in the goal area, as an
instance of \emph{Monitor Ball} is still at the back of the queue.

\section{The Behavior Control Framework}
\label{chap:bcf}

\subsection{Overview}
\label{sec:bcfoverview}

\begin{figure}
\centering
\includegraphics{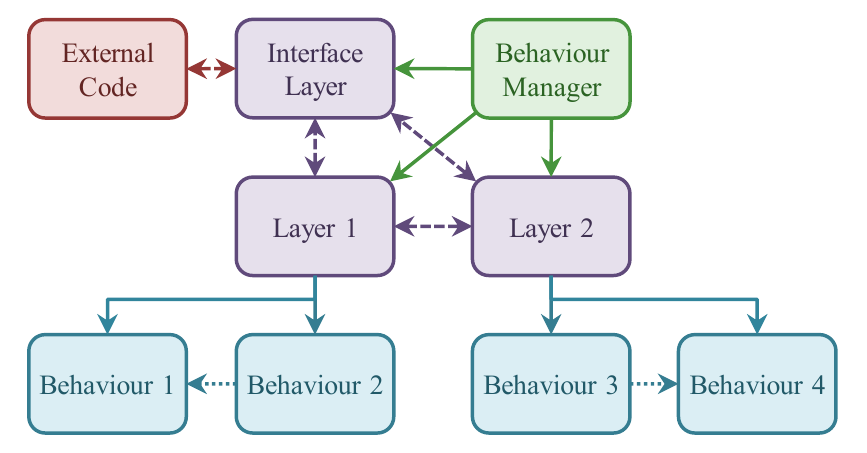}
\caption{A block diagram of the \bcfs architecture. Solid arrows indicate the %
object hierarchy, dashed arrows indicate the data exchange interfaces, and
dotted arrows indicate sample inhibitions.}
\label{fig:bcfarch}
\end{figure}

The \bcf \cite{BCFSourceforge} is a generic platform independent \cpp framework designed for behavior
control on robotic platforms. It is intended for the implementation of mid- to
high complexity agent behaviors. The main idea behind the framework is to
separate the control task into a pool of independent behaviors, partitioned
into so-called \emph{behavior layers}, where each behavior can be defined to
inhibit any number of other behaviors from within the same layer. The layers
are generally organized in a total order of decreasing abstraction and
resolution, and share information via virtual actuators and sensors, controlled
by corresponding \emph{actuator and sensor managers}. A parent \emph{behavior
manager} links all of the layers together and implements a step routine that
controls the execution of the entire structure. The layers are executed in a
user-defined order, generally corresponding to the total order from highest
level of abstraction to lowest level of abstraction. A key feature of the \bcfs
is that multiple behaviors can concurrently be activated in each layer. An
overview of the \bcfs architecture is presented in \figref{bcfarch}.

\subsection{Motivation}
\label{sec:bcfmotivation}

As discussed in \secref{sclmotivation}, the \bcf was developed as a more
powerful and complete solution to the behavior control problem for the \nop.
Being suitable for use in all application sizes down to the simplest of
controllers was no longer a requirement, as it was for the \scls. Instead, the
focus was on the creation of a framework that would facilitate the
implementation of complex behavior controllers, suitable for use on the \nop
and for humanoid soccer. Performance and efficiency of the framework were still
of high consideration, as well as its integrability and interoperability with
the remaining code. The \bcfs was inspired by, and based on, a custom behavior
control architecture that had been in development and use by team \nimbro for
almost a decade \cite{HierarchicalReactive}. Work actually started on the \bcfs
as an attempt to extract the architecture of this tried and tested custom
implementation into a standalone framework. In the process however, a number of
distinct changes were made in order to address the remaining weaknesses of the
architecture, while striving to retain its many strengths. Usability, structure
and customizability are examples aspects of the architecture to which
improvements were made.

\subsection{Behavior Inhibitions}
\label{sec:bcfinhibitions}

The inhibitions between the behaviors of each behavior layer are processed at
the beginning of program execution, before the step routine is first called. At
this point, the inhibition definitions are compiled into a directed acyclic
graph, referred to as the \emph{inhibition tree}. It is strictly an error if a
cycle in the inhibitions exist, as this would lead to unpredictable behavior
activations. Individual inhibition definitions can be specified as being either
\emph{chaining} or \emph{non-chaining}. The chaining inhibitions are considered
to act transitively with other chaining inhibitions, leading to additional
implicitly defined inhibitions, while the non-chaining inhibitions do not. Once
the inhibition tree has been established, the behaviors are topologically
sorted with respect to it, in order to ensure that the resolution of the
inhibitions at runtime is unambiguous.

At the beginning of every step, each behavior in a layer is queried for its
requested activation level. This is a real number on the unit interval and is a
measure of how relevant a behavior is to the current perceived situation. A
value of $1.0$ corresponds to a request for complete activation, while $0.0$
corresponds to complete deactivation. The activation levels are used for two
purposes, to evaluate which behavior(s) are active in a layer at any one time,
and to aggregate actuator values, as discussed in \secref{bcfcommunication}. The
behaviors are traversed in their topological order, and the respective
inhibitions are applied multiplicatively. This means for example that if a
behavior with an activation level of $0.7$ inhibits another behavior of
activation level $0.9$, then the latter will have its activation level reduced through
multiplication by $1-0.7=0.3$, to $0.27$. In by far the most common case, this means that a
behavior with an activation level of $1.0$ completely prevents all of the
behaviors it inhibits from executing. In this way, the requested activation
levels are refined into a set of true activation levels.

\subsection{Behavior Layer Data Interfaces}
\label{sec:bcfcommunication}

As the hierarchy of behavior layers are executed during a step from the top
down, it is generally required that the output of higher order planning in the
upper layers is made available to the lower layers. This is done using a network
of virtual actuators and sensors. Each layer receives data through its sensors
and delivers its output via its actuators. This is a single sender multiple
receiver arrangement, where multiple sensors in multiple layers can request to
receive the data from the same actuator. Actuators are uniquely identified by
name, and support the use of arbitrary data types for information exchange. If
the data type numerically supports it, an actuator can be made to be
aggregatable. This allows multiple concurrently active behaviors to write to the
same actuator. The output that is read by the corresponding sensors is then
calculated as the average of the written values, weighted by activation level.
This allows competing behaviors to have combined influence on an agent, provided
this is desired.

In addition to the transfer of data between layers, there is usually also a need
to exchange data with external sources. Most commonly this is in the form of
real-world sensory perceptions and motion commands. The concept of interface
layers exists for this purpose. From the perspective of the behavior manager,
this is simply a normal behavior layer with a slightly modified time of
callback execution. This is necessary so that the external data can be sent and
received at the appropriate times within a step. In the case of the \nop robot,
a ROS interface layer was implemented to allow communication of the behaviors
node with the other nodes in the system via the inbuilt ROS topics and services.
Interface layers also make it possible to split up a behavior control system
over process boundaries, meaning that multiple loop rates can be used. For
example, higher layers can be made to execute at a slower rate than the
more time-critical lower layers.

\subsection{Example}
\label{sec:bcfexample}

\begin{figure}
\centering
\includegraphics{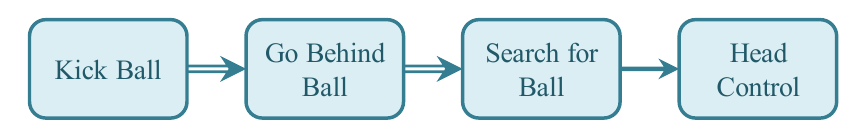}
\caption{A simplified example of a ball approach and kick behavior using the
\bcfs. Boxes represent behaviors, and the double and single arrows between
them represent chaining and non-chaining inhibitions respectively.}
\label{fig:walkandkick}
\end{figure}

Consider the simple ball approach and kick behavior presented in
\figref{walkandkick}. For simplicity, the system has been implemented
using only a single behavior layer. The blocks represent behaviors, and the
double and single arrows between them represent the chaining and non-chaining
inhibitions respectively. It should be noted that although no direct arrow
exists between \emph{Kick Ball} and \emph{Search for Ball}, the former still
inhibits the latter implicitly via the \emph{Go Behind Ball} behavior due to
the effect of chaining. Only \emph{Search for Ball} inhibits \emph{Head Control}
however, either explicitly or implicitly. In this arrangement each behavior
would return a requested activation level of 1.0 if the suitable preconditions
are met, and 0.0 otherwise. For example, \emph{Go Behind Ball} would return a
1.0 whenever the sensory perceptions indicate that the ball can be seen on the
field, and \emph{Search for Ball} would always return 1.0. Note however that the
\emph{Head Control} behavior can still execute whenever the \emph{Search for
Ball} behavior is itself inhibited.

A strength of the use of inhibitions is that it leads to a behavioral switching
dynamics that is somewhat automated, functioning without the explicit definition
of any transitions or the like. For example, consider the \emph{Kick Ball}
behavior, which returns an activation level of 1.0 if and only if the ball is
directly in front of a foot. Assuming the ball is visible, the robot would walk
towards the ball until the preconditions of the kick are met, at which point
\emph{Kick Ball} would automatically activate and suppress the \emph{Go Behind
Ball} walking behavior. Furthermore, on conclusion of the kick, the walking
behavior would automatically reactivate as soon as the kicking behavior
reports a zero activation level again. If the ball is suddenly observed once more in
an appropriate kicking location, then the walking behavior, or for that matter
whichever other behavior is running at the time, once again would temporarily
be suppressed while the kick takes over. The advantage of using inhibitions is
that all these transitions, which would normally explicitly have to be
specified, can be summarized into a very select few inhibition definitions.

\section{Conclusion}
\label{chap:conclusion}

Two frameworks for behavior control have been presented in this paper, the \scl
and the \bcf. The former is an effective framework for implementing low to
mid-level complexity agent behaviors, especially behaviors that have a
tendency of requiring structured sequences of actions. It has been
demonstrated to be useful in implementing miscellaneous finite state machines
required throughout the \nop platform code. One limitation of the framework
however, is that it does not inherently allow more than one basic behavior to be
active at any one time, even if this effect could theoretically be achieved with
the use of multithreading. This point, amongst many others, was addressed by the
advent of the \bcfs, which utilizes a tree of behavior inhibitions to evaluate
at every instant in time which behavior(s) should be activated. This allows
multiple aspects of an agent to be controlled simultaneously by independent
behaviors. The \bcfs was intended for the implementation of agent behaviors
from mid- to high complexity, more so than the \scls, but it does not preempt
the use of the latter. For instance, the \scls can still be used within the
individual behaviors of a \bcfs architecture to implement action sequences
based on finite state machines. Both frameworks were designed with performance and
efficiency in mind, and form a robust base on which a behavior control system
can be built.

\section*{Acknowledgement}

This work was partially funded by grant BE 2556/10 of the German Research
Foundation (DFG).

\bibliographystyle{IEEEtran}
\bibliography{ms}

\begin{thebibliography}{10}
\providecommand{\url}[1]{#1}
\csname url@samestyle\endcsname
\providecommand{\newblock}{\relax}
\providecommand{\bibinfo}[2]{#2}
\providecommand{\BIBentrySTDinterwordspacing}{\spaceskip=0pt\relax}
\providecommand{\BIBentryALTinterwordstretchfactor}{4}
\providecommand{\BIBentryALTinterwordspacing}{\spaceskip=\fontdimen2\font plus
\BIBentryALTinterwordstretchfactor\fontdimen3\font minus
  \fontdimen4\font\relax}
\providecommand{\BIBforeignlanguage}[2]{{%
\expandafter\ifx\csname l@#1\endcsname\relax
\typeout{** WARNING: IEEEtran.bst: No hyphenation pattern has been}%
\typeout{** loaded for the language `#1'. Using the pattern for}%
\typeout{** the default language instead.}%
\else
\language=\csname l@#1\endcsname
\fi
#2}}
\providecommand{\BIBdecl}{\relax}
\BIBdecl

\bibitem{BehaviourDefn}
R.~Wallace, G.~P. Sanders, and R.~J. Ferl, \emph{Biology: The Science of Life},
  3rd~ed.\hskip 1em plus 0.5em minus 0.4em\relax New York: Harper Collins,
  1992.

\bibitem{NimbroOPDescription13}
M.~Schwarz, J.~Pastrana, P.~Allgeuer, M.~Schreiber, S.~Schueller, M.~Missura,
  and S.~Behnke, ``{Humanoid TeenSize Open Platform NimbRo-OP},'' in
  \emph{Proceedings of 17th RoboCup International Symposium}, Eindhoven,
  Netherlands, 2013.

\bibitem{ArtificialIntelligence}
S.~Russell and P.~Norvig, \emph{Artificial {I}ntelligence: {A} {M}odern
  {A}pproach}.\hskip 1em plus 0.5em minus 0.4em\relax New Jersey: Prentice
  Hall, 1995.

\bibitem{Rao91modelingrational}
A.~S. Rao and M.~P. Georgeff, ``Modeling rational agents within a
  {BDI}-architecture,'' in \emph{Proceedings of the Second International
  Conference on Principles of Knowledge Representation and Reasoning
  (KR91)}.\hskip 1em plus 0.5em minus 0.4em\relax Morgan Kaufmann, 1991, pp.
  473--484.

\bibitem{Rao95bdiagents}
A.~S. Rao and M.~P. Georgeff, ``{BDI} agents: {F}rom theory to practice,'' in
  \emph{Proceedings of the First International Conference on Multi-Agent
  Systems (ICMAS-95)}.\hskip 1em plus 0.5em minus 0.4em\relax AAAI Press, 1995,
  pp. 312--319.

\bibitem{Rao95formalmodels}
A.~S. Rao and M.~P. Georgeff, ``Formal models and decision procedures for
  multi-agent systems,'' {Technical Note}, 1995.

\bibitem{Burkhard1998}
H.~D. Burkhard, M.~Hannebauer, and J.~Wendler, ``{Belief-Desire-Intention}
  deliberation in artificial soccer,'' \emph{AI Magazine}, vol.~19, no.~3, pp.
  87--93, 1998.

\bibitem{Brooks86Orig}
R.~Brooks, ``A robust layered control system for a mobile robot,'' \emph{IEEE
  Journal of Robotics and Automation}, vol.~2, no.~1, pp. 14--23, 1986.

\bibitem{Brooks89}
R.~Brooks, ``A robot that walks; {E}mergent behaviors from a carefully evolved
  network,'' \emph{Neural Comput.}, vol.~1, no.~2, pp. 253--262, 1989.

\bibitem{Connell89}
J.~Connell, ``A colony architecture for an artificial creature,'' Ph.D.
  dissertation, Electrical Engineering and Computer Science, May 1989.

\bibitem{Brooks91Ex}
R.~Brooks, ``Intelligence without representation,'' \emph{Artificial
  Intelligence}, vol.~47, pp. 139--159, 1991.

\bibitem{Maes1991ANA}
P.~Maes, ``The agent network architecture {(ANA)},'' \emph{SIGART Bulletin},
  vol.~2, no.~4, pp. 115--120, 1991.

\bibitem{Maes1990Sit}
P.~Maes, ``Situated agents can have goals,'' \emph{Robotics and Autonomous
  Systems}, vol.~6, no. 1–2, pp. 49--70, 1990.

\bibitem{BehaviourLanguage}
R.~Brooks, ``The {B}ehaviour {L}anguage; {U}ser's {G}uide,'' MIT AI Lab, 1990.

\bibitem{MacKenzie1997Thesis}
D.~MacKenzie, ``A design methodology for the configuration of behavior-based
  mobile robots,'' Ph.D. dissertation, Georgia Institute of Technology, GA,
  USA, 1997.

\bibitem{Colbert97}
K.~Konolige, ``Colbert: {A} language for reactive control in {S}apphira,'' in
  \emph{KI-97: Advances in Artificial Intelligence}, ser. Lecture Notes in
  Computer Science.\hskip 1em plus 0.5em minus 0.4em\relax Springer Berlin
  Heidelberg, 1997, pp. 31--52.

\bibitem{Saphira98}
K.~Konolige and K.~Myers, ``The {S}aphira architecture for autonomous mobile
  robots,'' in \emph{Artificial intelligence and mobile robots}.\hskip 1em plus
  0.5em minus 0.4em\relax Cambridge, MA, USA: MIT Press, 1998, ch.~9, pp.
  211--242.

\bibitem{XABSLThesis04}
M.~L{\"o}tzsch, ``{XABSL} - {A} behavior engineering system for autonomous
  agents,'' Diploma thesis. Humboldt-Universität zu Berlin, 2004.

\bibitem{XABSLThesis09}
\BIBentryALTinterwordspacing
M.~Risler, ``Behavior control for single and multiple autonomous agents based
  on hierarchical finite state machines,'' Fortschritt-Berichte VDI, Technische
  Universität Darmstadt, May 15 2009. [Online]. Available:
  \url{http://tuprints.ulb.tu-darmstadt.de/2046}
\BIBentrySTDinterwordspacing

\bibitem{hierarchicalFSMs}
A.~Kurt and {\"U}.~{\"O}zg{\"u}ner, ``Hierarchical finite state machines for
  autonomous mobile systems,'' \emph{Control Engineering Practice}, vol.~21,
  no.~2, pp. 184--194, 2013.

\bibitem{MachineObjects}
\BIBentryALTinterwordspacing
E.~Hiti. (2005) {The Machine Objects Class Library}. [Online]. Available:
  \url{http://ehiti.de/machine_objects/}
\BIBentrySTDinterwordspacing

\bibitem{SCLSourceforge}
\BIBentryALTinterwordspacing
P.~Allgeuer. (2013, Jul) {State Controller Library}. [Online]. Available:
  \url{http://sourceforge.net/projects/statecontroller/}
\BIBentrySTDinterwordspacing

\bibitem{BCFSourceforge}
\BIBentryALTinterwordspacing
P.~Allgeuer. (2013, Sep) {Behaviour Control Framework}. [Online]. Available:
  \url{http://sourceforge.net/projects/behaviourcontrol/}
\BIBentrySTDinterwordspacing

\bibitem{HierarchicalReactive}
S.~Behnke and J.~St\"uckler, ``Hierarchical reactive control for humanoid
  soccer robots,'' \emph{International Journal of Humanoid Robots (IJHR)},
  vol.~5, no.~3, pp. 375--396, 2008.

\end{thebibliography}

\end{document}